\documentclass[conference]{ieeeconf}
\IEEEoverridecommandlockouts
\usepackage{cite}
\usepackage{amsmath,amssymb,amsfonts}
\usepackage{algorithm, algorithmic}
\usepackage{graphicx}
\usepackage{dsfont}
\usepackage{textcomp}
\usepackage{xcolor}
\usepackage{hyperref}
\usepackage{wrapfig}

\def\BibTeX{{\rm B\kern-.05em{\sc i\kern-.025em b}\kern-.08em
    T\kern-.1667em\lower.7ex\hbox{E}\kern-.125emX}}
    
\newcommand\blfootnote[1]{%
  \begingroup
  \renewcommand\thefootnote{}\footnote{#1}%
  \addtocounter{footnote}{-1}%
  \endgroup
}

\title{
TRASS: Time Reversal as Self-Supervision 
}
\author{
\textbf{Suraj Nair$^{1, \dagger}$, Mohammad Babaeizadeh$^2$, Chelsea Finn$^{1,2}$, Sergey Levine$^2$, Vikash Kumar$^2$}\\
\texttt{surajn@stanford.edu,   \{mbz,chelseaf,slevine,vikashplus\}@google.com}\\
\authorblockA{\textit{$^1$Stanford University, $^2$Robotics at Google}}
}
    
\begin{document}

\maketitle

\begin{abstract}
A longstanding challenge in robot learning for manipulation tasks has been the ability to generalize to varying initial conditions, diverse objects, and changing objectives. Learning based approaches have shown promise in producing robust policies, but require heavy supervision and large number of environment interactions, especially from visual inputs. We propose a novel self-supervision technique that uses time-reversal to provide high level supervision to reach goals. In particular, we introduce the time-reversal model (TRM), a self-supervised model which explores outward from a set of goal states and learns to predict these trajectories in reverse. This provides a high level plan towards goals, allowing us to learn complex manipulation tasks with no demonstrations or exploration at test time. We test our method on the domain of assembly, specifically the mating of tetris-style block pairs. Using our method operating atop visual model predictive control, we are able to assemble tetris blocks on a KuKa IIWA-7
using only uncalibrated RGB camera input, and generalize to unseen block pairs.
Project's-page: \url{https://sites.google.com/view/time-reversal}
\end{abstract}

% \begin{IEEEkeywords}
% component, formatting, style, styling, insert
% \end{IEEEkeywords}
\section{Introduction}

\blfootnote{$^\dagger$Work completed at Robotics at Google}
Learning general policies for complex manipulation tasks often requires being robust to unseen objects and noisy scenes. 
However, learning complex tasks, in particular from visual inputs, present a number of challenges: (1) efficiently exploring the state space, (2) acquiring the suitable visual representation for the task, and (3) learning to execute fine control from dense input. To combat these issues, many methods rely heavily on some form of supervision, either as demonstrations, shaped rewards, or privileged state information. However, acquiring such supervision can be very costly. Shaped rewards often require significant tuning, demonstrations require experts to complete the task many times as well as complex recording infrastructure, and acquiring privileged state information often makes strong assumptions about the environment.
As an alternative to relying on outside supervision, we ask the question: ``Can an autonomous agent acquire this supervision on its own?'' In attempting to answer this question, our critical insight is that for many manipulation tasks, solving the task directly is difficult, while inverting it (changing the scene from solved to un-solved) is easy. To that end, we propose a novel method for gaining self-supervision that operates by exploring outward from a set of goal states, and \textit{reversing} the trajectories. We train a \textbf{time-reversal model~(TRM)} to predict these reversed trajectories, thereby creating a source of supervision leading to the goal. 

\begin{figure}[t]
\centerline{\includegraphics[width=0.99\columnwidth]{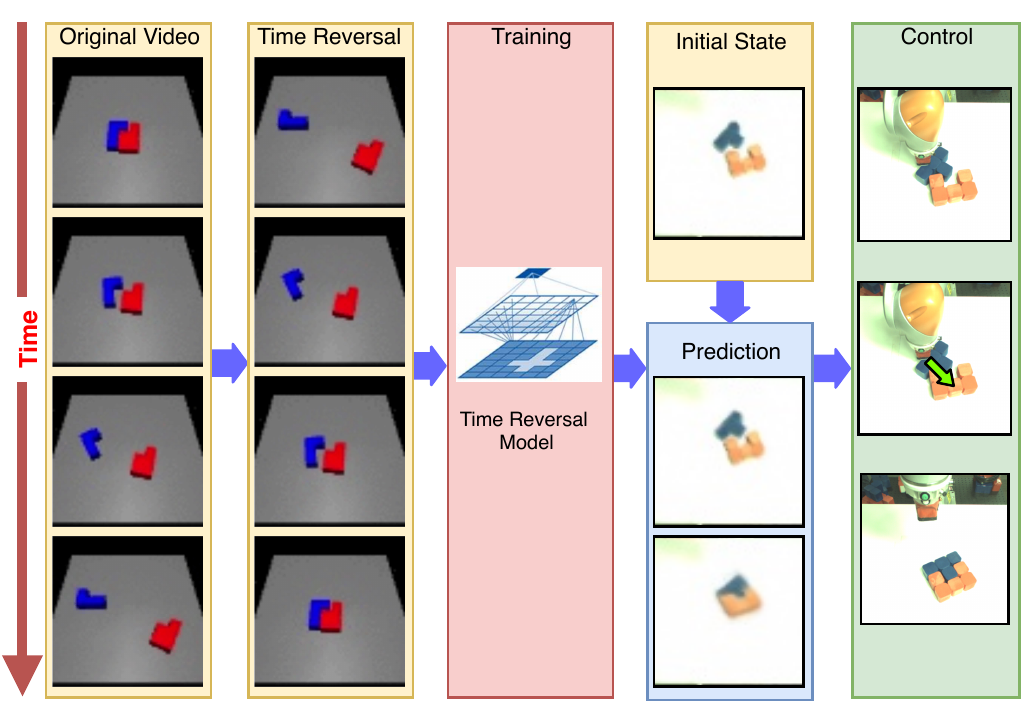}}
\vspace{-0.3cm}
\caption{\textbf{Time Reversal as Self-Supervision:} Our method (1) collects trajectories exploring outward from a set of goal states and reverses them. It then (2) trains a supervised model, the time reversal model (TRM), to predict these reversed trajectories. (3) It uses the trained TRM to predict the trajectory of states leading to the goal state for a new scene. (4) Control is executed to follow the trajectory predicted by TRM. }
\label{summaryfig}
\vspace{-0.6cm}
\end{figure}

Most manipulation tasks that one would want to solve require some understanding of objects and how they interact. However understanding object relationships in a task specific context is non-trivial. Consider the task of putting a cap on a pen. Successful task completion is dependent on both concentric alignment of axis and a particular approach direction. Thus we argue that learning the schematic understanding of objects and their relationships for manipulation requires (1) contextual understanding, (2) high level reasoning, and (3) precise control. 
However, uncapping the pen is a relatively simpler task, requiring less exploration and fine control to achieve. Yet, by uncapping the pen, one can learn the contextual understanding and high level reasoning needed to re-cap it. This is the key principle behind our time-reversal method. TRM models trained on reversed uncapping trajectories could provide strong supervision about coaxial alignment and approach direction. TRM's supervision can then be consumed by any local planner for execution.

The time-reversal method works by first collecting data by initializing to some set of goal states, applying random forces to disrupt the scene, and recording the subsequent states. We consider this self supervised, as the policy which disrupts the scene can just be random motion as it is in our setting, and thus does not require human input. We then train the TRM to predict these trajectories in reverse. At test time (when the goal states are unknown), the trained model can take the current state as input and provide supervision towards the goal, in the form of a guiding trajectory of frames leading from current state towards the goal (See Figure \ref{summaryfig}).
This guiding trajectory can be used as indirect supervision to generate a low level control policy via any model based or model-free techniques. Additionally, the time-reversal method can operate in any state domain, such as robot joint states, latent states, or raw images.
Note that the time-reversal model does not use or predict actions, purely state trajectories. This circumvents the issue of many actions being irreversible, and decouples high level task reasoning from low level control, each of which can be learned separately. This decoupling of high level planning and low level control should make each low level control step less planning intensive, thus enabling more complex, multi-stage tasks.

We test our method on the task of Tetris block mating and attempt to learn the semantic understanding necessary to solve the problem from raw visual inputs. Tetris (or Lego) block mating, owing to its close resemblance with industrial assembly applications, has been studied in the past \cite{haarnojacomposable} as it nicely casts the challenges of assembly in a set up that can be closely studied. These tasks require high level reasoning of how pieces must fit together, as well as fine control necessary to actually fit them together, making them especially challenging for existing methods. 
% In our results, we show that TRM outperforms model-based reinforcement learning like in \cite{DBLP:journals/corr/FinnL16} with a shaped cost, while using significantly less supervision.
% Additionally, a state of the art model-free method, D4PG \cite{d4pg} fails to learn when the reward is highly sparse, and even when it does learn fails to generalize to unseen block pairs, despite using privileged state information instead of images.
Experimental results show that TRM can reliably provide supervision towards the goal configuration entirely in visual space, and by using TRM with visual model predictive control (MPC), we achieve a success rate more than double that of the more heavily supervised visual MPC baseline, while achieving comparable performance to using ground truth full supervision.

In addition, we show that using TRM trained with data generated in a MuJoCo simulation \cite{mujoco} with domain randomization \cite{DBLP:journals/corr/SadeghiL16}, we are able to perform sim-to-real transfer, we are able to achieve a 75\% success rate of block pair mating on a real robot setup using a KUKA IIWA and only uncalibrated RGB camera input (No information between camera frame and robot frame is provided). 

\textbf{Summary of Contributions:} (1) Our primary contribution is a novel method for self-supervision that uses time-reversal to predict guiding trajectories towards goals. (2) We demonstrate that this method can be used in conjunction with a control policy to execute tasks, with higher success rates than more heavily supervised methods. (3) We show that this method can enable completion of tasks on real robots using training in simulation with domain randomization.

\section{Related Work}

Methods for robot control from visual inputs have been demonstrated on problems ranging from driving~\cite{NIPS1988_95} to soccer~\cite{rlsoccer}. One approach has been visual servoing, which directly performs closed loop control on image features~\cite{143350,6907309,538974}. While some visual servoing methods work with uncalibrated camera inputs ~\cite{606723,350995}, the general hand crafted nature of visual servoing limits the complexity of visual inputs and tasks where it can be applied. Other works have emphasized learning based approaches, in particular the use of deep neural networks to learn visuo-motor control from images~\cite{NIPS1988_95,hadsell,6252823,rlsoccer}.
These methods have shown impressive results in the problems of simple manipulation and grasping~\cite{DBLP:journals/corr/LevineFDA15,DBLP:journals/corr/PintoG15,DBLP:journals/corr/LevinePKQ16, DBLP:journals/corr/GhadirzadehMKB17,DBLP:journals/corr/MahlerLNLDLOG17,qtopt} by learning task specific policies. While these approaches are generally successful in their task, they have not been demonstrated on more complex tasks, particularly those which require both high-level planning and precise control. These methods have also been shown to work well when trained in simulation with domain randomization and transferred to a physical robot \cite{DBLP:journals/corr/SadeghiL16,DBLP:journals/corr/JamesDJ17,DBLP:journals/corr/TobinFRSZA17}
, a strategy which we use in our method.

Model based approaches to robot control have traditionally been most effective in tasks with low-dimensional states, such as helicopter control~\cite{helicopter} and robotic cutting~\cite{Lenz2015DeepMPCLD}, however recent methods have found success in learning a dynamics model in image space~\cite{DBLP:journals/corr/FinnL16,DBLP:journals/corr/abs-1710-11311, DBLP:journals/corr/abs-1710-05268}. 
% Similarly, Agarwal et al.~\cite{DBLP:journals/corr/AgrawalNAML16} and Nair et al.~\cite{DBLP:journals/corr/NairCAIAML17} have learned inverse dynamics models in image space. 
These models have been shown to be effective in planning~\cite{DBLP:journals/corr/abs-1804-00062}, and have even been extended to operate in 3D point cloud space~\cite{DBLP:journals/corr/ByravanF16}. While these approaches work well on simple tasks, they require additional information during evaluation in the form of either goal images or demonstrations, exactly what our method circumvents.

At the same time, exploration of visual domains remains a significant challenge. A number of recent works have tried to tackle this problem in low dimensional spaces by training goal-conditioned policies, and reformulating seen states as goals as self-supervision, yielding improved sample efficiency~\cite{DBLP:journals/corr/AndrychowiczWRS17,DBLP:journals/corr/abs-1802-09081}. A similar idea has been extended to physical robots and images by Nair and Pong et al~\cite{imaginedgoals}, who practice reaching imagined goals. This method however still requires goal images at test-time, and tests on a simpler puck-pushing tasks.

Another approach to self-supervised exploration involves resetting to goal states and exploring states around the goal state ~\cite{DBLP:journals/corr/abs-1803-10227,DBLP:journals/corr/FlorensaHWA17,DBLP:journals/corr/abs-1804-00379}. While these methods are most similar to our approach, they still rely on supervision in the form of rewards from the environment. As a result these methods still need exploration unlike our method which needs neither goal specification nor exploration at evaluation time. These approaches also have not been shown on physical robots with image input.

\section{Preliminaries}

We formulate the space of problems in which our method can be applied as finite-horizon, Markov decision processes with sparse rewards. At each timestep $t$, the agent receives a state $s_t \in S$ and chooses an action $a_t \in A$ based on a stochastic policy $a_t \sim \pi(s_t)$. After taking an action the environment returns the stochastic next state $s_{t+1}$ and reward $r(s_t, a_t) =\mathds{1} \{s_{t+1} \in S_g\}$ where $S_g$ is a subset of $S$ consisting of all goal states.

Our method is well suited when: \textbf{(1)} During a training phase we can reset to some subset of goal states $S_g' \in S_g$, selected at random. \textbf{(2)} The Markov chain produced by taking uniformly random actions from any goal state $s_g \in S_g' $ has a non-zero probability of reaching all states $s \in S$. \textbf{(3)} If there exists transitions from states $s_i$ to $s_j$, then there exists some set of actions $A^*$ which can traverse from $s_j$ to $s_i$. Assumption 1 enables us to reset to goal states during the training phase. However, it does not provide any general specification of goal states nor any information about how to reach them. Furthermore, assumption 2 simply ensures that all states $s \in S$ can be reached from the goal, a condition satisfied in many robotics problems, including manipulation. Assumption 3 ensures that it is feasible to follow reversed trajectories. Our formulation does not assume that at evaluation time the objective is to reach a specific state $s_g$, but rather to reach any state $s_g \in S_g$, where no specification of the goal is provided. Thus a successful method must be able to (1) reason about what the goal state is given the current state and (2) execute control to reach it. 

Below we list some of the tasks where TRASS can be applied, and some where it cannot, with justification. In principle TRASS \textbf{can} be applied to:

\textit{2D assembly:} as demonstrated in our results.

\textit{3D assembly:} consider the task of block tower stacking \cite{overcomingexp_nair, yuke_diversevisuomotor}. Exploration of the robot from the goal state will knock the blocks down. In our results we observe that the time-reversal model (TRM) predicts the blocks floating through the air and reforming a tower given a new initialization. Note that while these videos are not perfect, i.e. they have multiple blocks moving simultaneously, this still provides a dense cost function which corresponds to task progress.

\textit{Tabletop/Desk Organization:} For example - turning on a computer or opening a drawer \cite{lynch2019play}. The state trajectory seen by releasing and moving away from a power button when reversed provides a strong supervision signal of reaching towards and pressing a button. Similarly, when starting at the goal state of gripping an open drawer, self-supervised exploration outward will either explore outward from the handle or push the drawer closed. Then the time reversal model will learn to predict trajectories of approaching the handle and pulling open the drawer.

\textit{Household Manipulation:} (1) Setting up a dining table \cite{ntp}, (2) Making a bed \cite{seita_bedmake_2019}, (3) Cleaning a room. (1) Starting from a properly set table, exploration will knock the plates/utensils around on the table. This trajectory reversed provides a useful cost signal in re-setting the table. (2) Starting from a made bed, random perturbations to the bed can crumple the blanket, which when reversed provides supervision on how to flatten and spread the blanket. (3) Similarly, randomly perturbing objects in a clean/organized room will distribute the objects around the room. These trajectories reversed will show objects being placed back to their correct positions, strong supervision for room cleaning.

TRASS \textbf{cannot} be applied to:

\textit{Gripper in state:} The primary situation where TRASS is not applicable is when (1) the gripper that manipulates objects is visible in state and (2) the dynamics between the gripper and objects are irreversible. For example, when grasping an object and holding it up, if the gripper is visible in the state, then there is no way to reverse the state trajectory (keep the gripper fixed in the air, while the block floats up into it). However - if the gripper is not part of the state, then the time-reversal trajectory will simply show the object moving up, which can provide a useful cost signal for the task, and can feasibly be followed by a policy. One way to avoid having the gripper in the state space is to use a high level action space, for example pick-place locations, or start-end push locations. There are some cases where even when the gripper is in the state, the dynamics are reversible and TRASS can be applied, for example turning a knob or closing a drawer.

\textit{Irreversible Object Deformations:} In cases where object deformations can be reversed like cloth folding \cite{matas_sim2real} or rope manipulation \cite{causalinfogan} TRASS is applicable. However in cases where object deformations cannot be reversed, for example cracking an egg, mixing two ingredients, or welding parts together, TRASS cannot be applied.”

% We acknowledge that a goal-conditioned variant of our approach is feasible, however in this work we limit our experiments to problems with a broad definition of goal.

\section{Method} \label{m}

\begin{figure}[t]
\centering
\includegraphics[width=0.99\columnwidth]{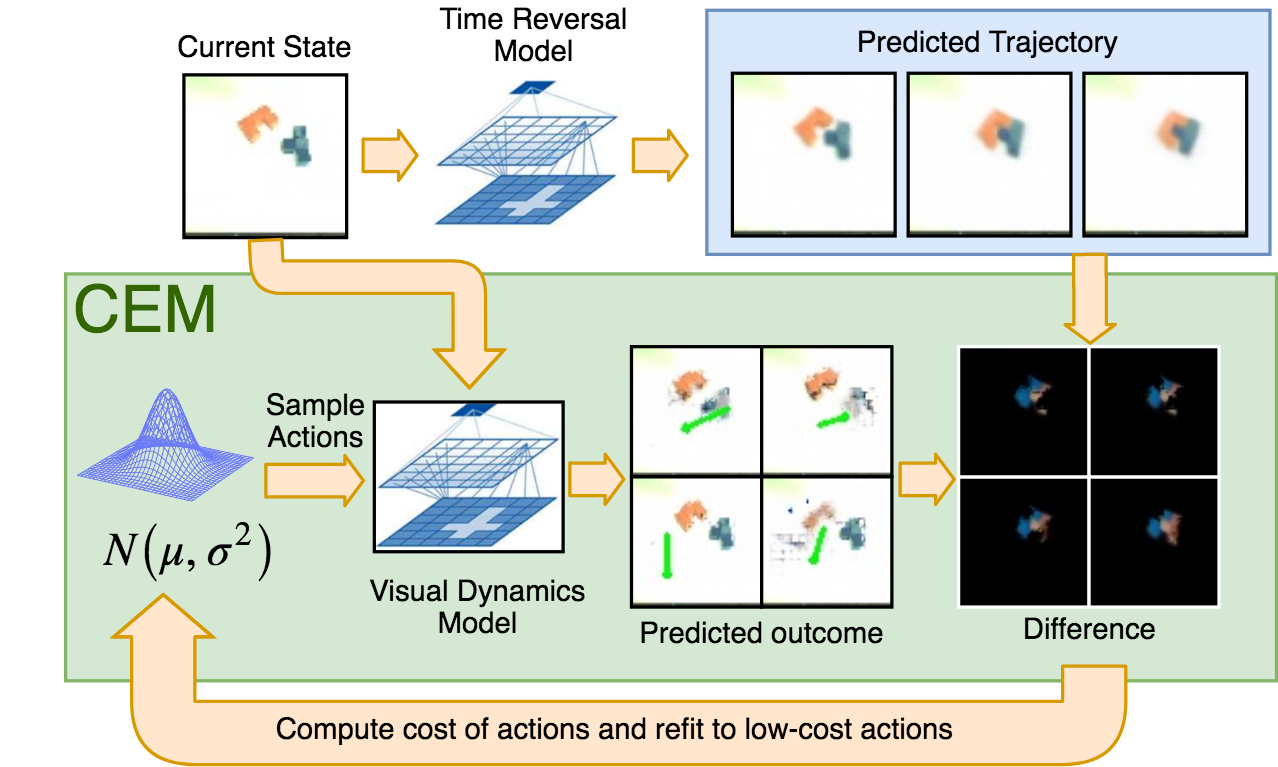}
\caption{\textbf{One step of TRASS:} A new image is provided and is fed through the time reversal model to produce a trajectory toward the goal state. Simultaneously, we use CEM to optimize actions with the cost of an action defined by its predicted next state similarity to the TRM trajectory.}
\label{inferencefig}
\vspace{-0.6cm}
\end{figure}

The core contribution of our method lies in acquiring supervision through time-reversal, which can be broken down into three distinct phases \textbf{(1)} Time-reversal exploration and data collection \textbf{(2)} Time-reversal model (TRM) training \textbf{(3)} Test time goal and trajectory prediction using time-reversal model. Then, using the goals and trajectories provided by the time reversal model, one can create a policy $a_t \sim \pi(s_t)$ to complete the task.

\textbf{Time-Reversal Exploration and Data Collection:} The motivation behind time-reversal is that for many tasks, it is much easier to invert the task, i.e. destroy the scene from a goal state, than to reach a goal state from an initial state. Thus, if the information collected by inverting the task can be used to reach goals, the need for external supervision to learn tasks can be reduced.

During the exploration and data collection phase, the scene resets to a goal state $s'_g \in S_g$, and the agent explores outward using random perturbations to the scene. Thus, using no expert knowledge, the agent collects a trajectory
$s'_g, s_{t+1}, s_{t+2}, ..., s_{t+M}$ which when reversed can be used by the agent to learn to return to the goal state.

For many manipulation tasks, a random exploration policy such as the one described is sufficient in exploring the space, in particular when randomly perturbing the scene the distribution of $s_{t+M}$ matches the distribution of initial states $s_0$. While this holds true for several domains of robotics (assembly, stacking, rearranging),  in some manipulation tasks this may not be the case. Consider the task of screwing a cap on a bottle. In this setting, random perturbations on the tightened cap are unlikely to yield states which represent the distribution of initial states (before the cap has been screwed on the bottle). However, we argue that our time reversal framework can be still be applied in these cases, using a different method of exploration\footnote{An exploration policy driven by curiosity in the form of state novelty could work well here. Left to future work.}.

\textbf{Time-Reversal Model Training:}
After the exploration and data collection phase, the agent has collected $k$ trajectories of the form $s_{ t}, s_{ t+1}, s_{ t+2}, ..., s_{ t+M}$ where $s_{t} \in S_g$ are potential goal states.

We then reverse these trajectories, and train the time-reversal model to predict a sequence of states along the reversed trajectories:
$$s^*_{t+M-1}, ...,s^*_{ t+M-(A-1)}, s^*_{ t+M-A} \sim TRM_{\theta}(s_{ t+M})$$
where $A$ is a tunable parameter for how many time steps time-reversal model predicts. The TRM is then trained to minimize the loss 
$L(TRM_\theta(s_{t+M}), s_{t+M-1}, ...,s_{t+M-(A-1)}, s_{t+M-A}])$ across batches of trajectories. 
% Details about the loss/architecture used can be found in the Supp. Material.

\textbf{Goal and Trajectory Prediction using TRM:}
At test time, the agent's objective is to reach some state in a new scene which satisfies the goal condition. In particular, the agent is initialized to some initial state $s_0$, and wants to reach some goal state $s'_g \in S_g$. The agent is not provided any supervision here besides knowing when the task is complete (i.e. there is no explicit reward function). Rather, the trained time reversal model predicts the sequence of states leading towards the goal state:
$$s^*_{t+1}, s^*_{t+2}, ..., s^*_{t+A} \sim TRM_{\theta}(s_t)$$ At every time step in the episode this trajectory is recomputed, producing a high level plan towards the goal state. We observe that TRM is able to directly predict sequence of the Tetris blocks moving together and into the mated configuration, all in image space (See Figure ~\ref{bwexs}).

Note that the time-reversal model \textbf{does not} predict actions, but rather trajectories of states. This decoupling serves two purposes. First, clearly many actions are not reversible, e.g. $p(s_{t+1} | s_t, a_t) \neq p(s_{t} | s_{t+1}, a_t) $, so learning the time reversal model with actions is not possible. Second, by allowing the TRM to focus exclusively on state trajectories, it is able to learn a high level plan purely in state space, decoupling the control from high level guidance. Then, using a trained time-reversal model, control can be learned separately. Details on the control method can be found in Section \ref{mdet}.

\section{Method Details} \label{mdet}
While the time-reversal framework is agnostic to the choice of state space, we examine the problem of Tetris block pair mating from raw RGB input. Thus our time-reversal model becomes a video prediction model. Specifically we use identical architecture and training parameters as \cite{sv2p}. Additionally, while the time-reversal framework can operate with any control policy, we use visual model predictive control as in \cite{DBLP:journals/corr/FinnL16}, the details of which are provided below.

\textbf{Using TRM Prediction for Control:} 
Given the time-reversal model's predictions  (trajectory leading towards the goal), an agent can use any number of control methods to take actions which follow this trajectory. For example, using model-based techniques, one can use a model to plan actions which follows the states predicted by the TRM. Using model-free approaches, one can learn a policy using a reward function which gives high reward for following a state trajectory similar to the TRM prediction. In this work, we use visual model predictive control as a means to take actions which follow the TRM trajectory  (See Figure \ref{inferencefig}), due to the ability to learn visual MPC from purely self-supervised data, and its ability to generalize to novel tasks and objects.

We learn the visual dynamics model through random actions as done in \cite{DBLP:journals/corr/FinnL16}. The visual dynamics model is also an instance of the SV2P model, and is trained using an identical architecture and loss to the time-reversal model. The only difference is that the visual dynamics model is trained on temporally  ordered sequences (not reversed), and includes actions. Given the current state, and a sequence of $N$ actions, the visual dynamics model is trained to predict the next $N$ states:
$$s^*_{t+1}, s^*_{t+2}, .., s^*_{t+N} \sim F_\delta(s_t, a_t, a_{t+1}, ... a_{t+N})$$

Then, given the trained visual dynamics model and time-reversal model, we plan actions using the cross entropy method (CEM) \cite{cem} such that the predicted future states align with the predicted TRM trajectory (See Figure \ref{inferencefig}).
Formally, this decomposes the control policy $a_t \sim \pi_{\theta, \delta}(s_t)$ into a visual dynamics model $F_{\delta}(s_t, a_t)$ and the time reversal model $TRM_{\theta}(s_t)$. 
$\pi_{\theta, \delta} = CEM(F_{\delta}, TRM_{\theta}, s_t) $

This policy $\pi_{\theta, \delta}$ is then used to receive states and produce actions at each step of evaluation. A visual depiction of one step of the policy is shown in Figure \ref{inferencefig}. At a single time step, the time-reversal model predicts the sequence of states leading towards a goal state, and iterative sampling and refitting is used to find actions for which the visual dynamics model predicts a trajectory which matches the time-reversal model trajectory.
% We initialize the Gaussian distribution for CEM to $\mathcal{N}(0, 0.1)$, sample 200 actions, refit to the 40 actions with lowest cost, and repeat at most 5 iterations before stepping the action with lowest cost in the environment. We use mean squared error between the predicted outcome images and TRM's 5th frame prediction as cost for each action.
Note that the learning of the visual dynamics model and use of CEM for planning is not novel and is simply a means to take actions which follow the time-reversal model trajectory. Our contribution lies in the time-reversal model itself, and any number of control methods can be used to plan with it. 
% Further details about CEM and planning parameters can be found in the supplementary material.

% \begin{figure}[t]
% \centerline{\includegraphics[width=\columnwidth]{figs/setupfigm.pdf}}
% \caption{Environments and Goals: \textit{top row}: the real setup consists of a KuKa IIWA arm mounted over a flat tabletop workspace. \textit{bottom row} the Mujoco simulation environment contains a similar setup of block pairs on a flat table top. The tool in the simulator is a single green block. \textit{left column}: initial states consist of the blocks randomly placed in the scene. \textit{right column}: Goal states consist of any state where the male and female part fit together to complete the 3x3 square. These can occur at any pose - the only criteria to be a goal state is that the parts are properly mated. }
% \label{obsgoalsfig}
% \end{figure}

\section{Experiment Details}

\textbf{Objects and Goals:}
Our problem of Tetris block mating consists of a set of blocks, decomposed, with each part randomly placed on a flat tabletop. A tool (simulated cube or a robot end-effector) is used to push the objects in the scene. Each block pair consists of two parts which when mated complete a 3x3 square.
The set of goal states $S_g$ consists of all states where the objects are combined to form the 3x3 square. Importantly, the goal configuration is pose invariant. As long as the 3x3 square is completed, the goal configuration is reached regardless of the location on the table or orientation where the completion occurred.  \textbf{Observation/Action Space:}
The observation space consists of 64x64 pixel angled top-down RGB images.
The action space is the high level action space $A \in R^4$ bounded from $-0.2$ to $0.2$, representing the start $<x,y>$ location and an end $<x,y>$ location of a push.
In practice, when an action is called with a start and end point, the robot endeffector (or simulated tool) moves to above the start location, moves down to the table height, pushes linearly towards the end point using Cartesian control with a force threshold, then lifts up and out of the scene, at which point the next state is captured.
\textbf{Environments:}
We primarily train our models on data generated in a simulation environment, and evaluate our methods on both the simulation environment and a real robot setup.
\underline{Simulation Environment:}
The simulation setup is built using the MuJoCo simulation engine \cite{mujoco}. It consists of a flat tabletop, upon which the male and female blocks can slide in the $x,y$ plane or rotate around the $z$ axis.The blocks are composed of several cubes (of side length 5cm), forming unique shapes. 
\underline{Robot Environment:}
The robot environment consists of a KUKA IIWA robot operating on a tabletop. The blocks dimensions are the same as in simulation.
% Details about the environment set up can be found in the supplementary material. 
% Additionally, the blocks in the physical world have less friction than in the simulation, making the physical setup less prone to random, unwanted rotations and forces, hence the task is slightly easier on the physical setup.
% \begin{figure}[t]
% \centerline{\includegraphics[width=0.95\columnwidth]{figs/envs.png}}
% \caption{Top: the real setup consists of a KuKa IIWA arm mounted over a flat tabletop workspace. Bottom: the Mujoco simulation environment contains a similar setup of block pairs on a flat table top. The tool in the simulator is a single green block.}
% \label{simrealfig}
% \end{figure}
% \begin{figure}[t]
% \centerline{\includegraphics[width=\columnwidth]{figs/setupfigm.pdf}}
% \caption{The Tetris Block Mating Task: \textit{top row}: the real setup consists of a KuKa IIWA arm mounted over a flat tabletop workspace. \textit{bottom row} the Mujoco simulation environment contains a similar setup of block pairs on a flat table top. The tool in the simulator is a single green block. \textit{left column}: initial states consist of the blocks randomly placed in the scene. \textit{right column}: Goal states consist of any state where the male and female part fit together to complete the 3x3 square. These can occur at any pose - the only criteria to be a goal state is that the parts are properly mated. }
% \label{obsgoalsfig}
% \end{figure}
\footnotetext{Note that since there are many valid goal states, the video prediction problem is highly stochastic. Hence, predicted images are likely to be blurrier than in more deterministic cases, such as in the visual dynamics model.}
\textbf{Data Collection:}
To train both the visual dynamics model and time reversal model, we primarily collect data in simulation. To transfer to the real world, and to improve overall performance, we apply domain randomization \cite{DBLP:journals/corr/SadeghiL16}. For the visual dynamics model we also explore collecting data on the real robot. 
% \begin{figure}[t]
% \centerline{\includegraphics[width=0.75\columnwidth]{figs/dr_2.png}}
% \caption{Domain Randomization: We randomize the color of the blocks, the position of the light, and the position/field of view of the camera to generate a randomized data set. We apply this randomization in the data for both the visual dynamics and time-reversal models, and it allows us to transfer these models a real robot setup without the need for fine-tuning on real data.}
% \label{dr}
% \end{figure}
\underline{Time-Reversal Model Data:}
% \begin{figure}[t]
% \centerline{\includegraphics[width=0.75\columnwidth]{figs/dr_2.png}}
% \caption{Domain Randomization: We randomize the color of the blocks, the position of the light, and the position/field of view of the camera to generate a randomized data set. We apply this randomization in the data for both the visual dynamics and time-reversal models, and it allows us to transfer these models a real robot setup without the need for fine-tuning on real data.}
% \label{dr}
% \end{figure}
To collect training data for the time reversal model, we collect \textit{dis-assembly} trajectories. Specifically, in a single trajectory we first initialize to a goal state $s_g \in S_g$ chosen uniformly at random. We then insert random forces to break apart the objects and record the subsequent trajectory of states $s_0 (s_g), s_1, ..., s_T$. We then save the trajectory in reverse: $s_T, ..., s_1, s_0 (s_g)$ \underline{Visual Dynamics Model Data:}
To collect training data for the visual dynamics model, we simply execute random actions, and capture the subsequent state. In a single trajectory, we initialize to a state $s_0 \in S$, sample actions uniformly, and save the transitions $(s_t, a_t, s_{t+1})$.

\section{Results}

\begin{figure}[t]
\centerline{
\includegraphics[width=0.99\columnwidth]{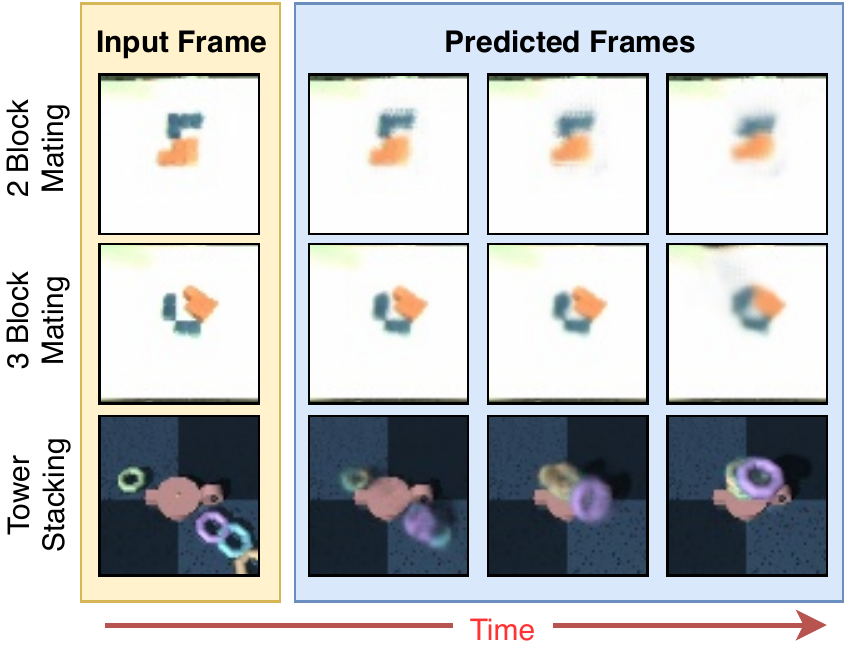}
}
\vspace{-0.3cm}
\caption{\textbf{Time Reversal Model Qualitative Examples:} Examples of the output of TRM from unseen initialization on the tasks of 2 Block Mating, 3 Block Mating, and Tower Stacking. In all cases the TRM trajectory provides a realistic plan to the goal.}
\label{bwexs}
\vspace{-0.6cm}
\end{figure}

Since the focus of our contribution is acquiring self-supervision through time reversal, in our experiments we ask the question \textit{``How does self-supervision from time-reversal compare to other more expensive forms of supervision?'}'
First, we examine qualitatively the supervision provided by TRM on 3 different tasks - 2 block assembly, 3 block assembly, and tower stacking in Figure \ref{bwexs}. We observe that in all cases TRM is able to provide a realistic trajectory leading to the goal from an unseen initialization.

Next, we compare the quantitative performance of our method against two other approaches with different levels of supervision on the 2 block assembly task:  (1) a \textbf{semi-supervised} approach which performs visual MPC with a shaped cost but no ground truth goal information and (2) a \textbf{fully supervised} approach which performs visual MPC with ground truth goal information (not available in real world, made available only in physics simulators). We observe that TRM based supervision outperforms the semi-supervised baseline and gets comparable performance to the full supervision.

\subsection{Baselines}

% \begin{figure}[t]
% \centerline{\includegraphics[width=1.0\columnwidth]{figs/ablation.png}}
% \caption{Ablation Study: We compare the ``Standard'' algorithm we use to other variants. The ``Standard'' algorithm uses TRM's 5th frame prediction in computing CEM cost, and does up to 5 steps of CEM, trains on Domain Randomized data, and does not use data from evaluation. We compare this with (1) using 10 frames from TRM instead of 5, (2) doing 10 steps of CEM, (3) doing 1 step of CEM, (4) Finetuning on 1000 trajectories from evaluation, and (5) without camera randomization in the domain randomization process. 500 trials each.}
% \label{ablations}
% \end{figure}

Below we describe the version of our method we use in our experiments and the two baselines which we compare our against. The success rates are shown in Figure \ref{baselines}.

% \textbf{D4PG \cite{d4pg}: Distributed Distributional Deep Deterministic Policy Gradient.} This model-free algorithm is trained for 1 million steps with a sparse reward $R(s) = 1$ if the blocks are mated and 0 otherwise. Unlike all of the other methods, D4PG uses ground truth object poses as the state space, not images, meaning that it requires privileged information. While it is able to succeed in the simplest setting (with a 91\% success rate), it completely fails when the reward is more sparse, and fails to generalize to unseen block pairs, unlike our method.

\textbf{TRM (Ours) (Self-Supervised):} Our method using  $CEM(F_{\delta}, TRM_{\theta}, s_t)$, as described in Section \ref{m}, using the 5th frame prediction from TRM to compute cost and doing max 5 iterations of CEM.

\textbf{Shaped Reward \cite{DBLP:journals/corr/FinnL16} (Semi-Supervised):} Performs visual model predictive control in the same way as in our method, using the same visual dynamics model. However, the cost of an action is computed as the area of the convex hull of the blocks in the predicted future image. The convex hull is computed by using color segmentation to identify the blocks, and solving for the convex hull in pixel space. The convex hull is a continuous signal that will be larger the farther away the blocks are, and will be minimized when the blocks are correctly mated, thus it provides a dense reward for the visual model predictive control. Despite using this shaped reward, this approach achieves a success rate roughly half of what our method achieves, indicating the value of explicitly predicting a trajectory toward the goal.

\textbf{Ground Truth Goal \cite{DBLP:journals/corr/FinnL16} (Fully-Supervised):}
In this setting, we perform visual model predictive control in the same way as in our method, using the same exact visual dynamics model. However, in this case, instead of planning actions for which the predicted future states match the TRM predicted trajectory, we plan actions for which the predicted future states match a ground truth goal image (generated by accessing the internals of the simulator). That is, at every time step of the episode the action comes from $CEM(F_{\delta}, s_g, s_t)$ using the difference to one goal image $s_g \in S_g$ to compute cost. This approach has a marginally higher success rate than our method (within 10\%), due to the fact that it uses the ground truth goal image, while our method learns to predict the goal. 

It is important to note that unlike our method, the Ground Truth Goal comparison cannot be extended to the real robot because it requires a ground truth goal image. Rather, it requires access to a simulator which can be used to compute and render the ground truth goal image. TRM (our method) does not use this information, but still achieves comparable performance (See Figure \ref{baselines}) by learning to predict the goal image and the trajectory towards it. TRM also outperforms the Semi-Supervised comparison with shaped cost.

\begin{figure}[t]
  \centering
  \begin{tabular}{|c||c|c||c|c|c|}
  \hline
  & Shaped  & TRM & Ground Truth \\ 
  & Reward \cite{DBLP:journals/corr/FinnL16}  & (Ours) & Goal \cite{DBLP:journals/corr/FinnL16} \\ 
%   & (Semi-Supervised)  & (Self-Supervised) & (Fully-Supervised) \\
  \hline\hline  % \hhline{|=|=|=|} 
   
  Seen  & $19 \pm 2$ & $49 \pm 2$  & $57 \pm 2$   \\ \hline
  
  Seen (Far) & $14 \pm 3$ & $56 \pm 2$  & $61 \pm 2$  \\ \hline
  
  Unseen & $26 \pm 2$ & $50 \pm 3$ & $55 \pm 2$  \\ \hline
 \end{tabular}
 
  \caption{\textbf{Success Rates on Simulated Tetris Block Mating}: Here we show the success rate of our time reversal method (self-supervised) compared to that of a shaped reward (semi-supervised), and ground truth goal information (fully-supervised), on three variants of the Tetris block mating task. In the \textit{Seen} variant, the models are tested on mating block pairs seen during training, in the \textit{Seen (Far)} variant, the models are tested on seen block pairs, but initialized farther away (at least 30 cm), and in the \textit{Unseen} variant, the models are tested on mating block pairs \textbf{not} seen during training. TRM significantly exceeds the performance of the shaped reward, while achieving comparable performance to the fully supervised approach.}
  \label{baselines}
  \vspace{-0.6cm}
\end{figure}

\subsection{Robot Results}

\begin{wrapfigure}{r}{0.4\linewidth}
 \includegraphics[width=0.4\columnwidth]{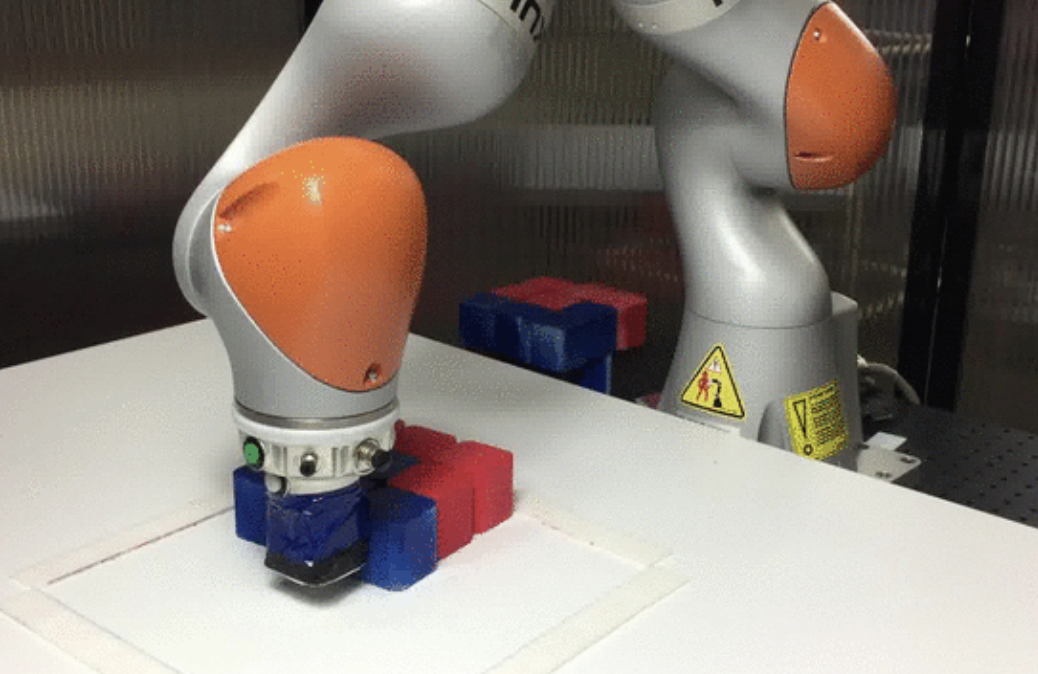}
 \caption{\textbf{Our Robot Setup:} KuKa IIWA-7 mounted over a flat tabletop.}
 \label{robotpic}
\end{wrapfigure}

We demonstrate that our approach successfully extends to a physical robot setup (See Figure \ref{robotpic}). In Figure \ref{robotresults} we report our method's performance on both seen and unseen blocks, as well as with fine-tuning on robot data and a modified version of domain randomization during training that has no camera randomization.  Our method successfully mates seen block pairs 75\% of the time and unseen block pairs 50\% of the time. We also find that fine tuning the visual dynamics model on the 825 robot trajectories and removing camera randomization has no significant impact. We suspect the lack of improvement from robot trajectories is due to benefits from aggressive domain randomization in training. Note: The success rates on the real robot are higher than those in simulation, due to the fact that the contact dynamics in simulation are much noisier, making mating more difficult.

 \begin{figure}
  \centering
  \vspace*{0.3cm}
  \begin{tabular}{|c|c||c|}
  \hline
  Training  & & Success  \\
  Data  &  Blocks  &  Rate \\ \hline\hline  % \hhline{|=|=|=|=|=|} 
   
  Standard & Seen & $75 \pm 10$  \\ \hline
  Standard & Unseen & $50 \pm 11$ \\ \hline
  + Robot Data &  Seen & $70 \pm 10$  \\ \hline
  Fixed Camera &  Seen & $75 \pm 10$  \\ \hline
 \end{tabular}
  \caption{\textbf{Success Rates on Real Robot:} We test transferring our method to a physical setup with a KuKa IIWA robot. We find that testing on seen blocks, it achieves 75\% success. We additionally report the performance on (1) unseen blocks, (2) finetuned on real robot pushing data, and (3) without camera randomization in the domain randomization process. 20 trials each.}
 
  \label{robotresults}
  \vspace{-0.6cm}

\end{figure}

\section{Limitations and Future Work}
The primary limitation of time-reversal is that it is restricted to settings with reachability between states, as described in Assumption 3 of the Preliminaries. While this is the case in a wide collection of tasks, there are some settings where it does not hold. Consider the example of cooking - once an egg has been cracked, there is no way it can be ``uncracked''. 

Another limitation of the current formulation is that we  explore tasks for which the abstract, high-level goals is fixed, while the explicit goal state for a given scene is unknown. While this may apply in simple assembly problems, more complex problems will have a broader space of goals. One direction of future work would be to modify the time-reversal model to accept some form of abstract goal specification, enabling goal conditioned TRM.
A limitation of the block mating task itself is that it is limited to the horizontal plane. While the multiple stages of the task make it difficult for state of the art methods to solve, the visual scenes are still fairly simple. We plan to extend this work to three dimensional structures, which will push the complexity of both time-reversal and control.
Another limitation is the nature of the time-reversal exploration. Currently, the exploration is done through random perturbations applied to the goal state, however in some complex tasks (such as screwing a cap on a bottle), this form of exploration is insufficient. Using exploration methods driven by state novelty is one possible approach to address this.
Lastly, blurriness in the video prediction has made it challenging to extend this work to more complex assembly problems with longer horizons, many objects, and complex degrees of freedom. One approach to address this would be to plan in latent space.

\section{Conclusion}
We have proposed a method which self-supervises task learning through time reversal. By exploring outward from a set of goal states and learning to predict these state trajectories in reverse, our method TRM is able to predict unknown goal states and the trajectory to reach them. This method used with visual model predictive control is capable of assembling Tetris style blocks with a physical robot using only visual inputs, while using no demonstrations or explicit supervision.
\color{black}

\section*{Acknowledgment}

We would like to thank Dumitru Erhan and others from Google Brain Video for valuable discussions. We would also like to thank Satoshi Kataoka, Kurt Konolige, Ken Oslund, Sherry Moore and others from Google Brain for help with experimental setup and infrastructure.

% \newpage

\bibliographystyle{unsrt}
\bibliography{trm}
% \begin{thebibliography}{00}
% \bibitem{b1} M. Andrychowicz, F. Wolski, A. Ray, J. Schneider, R. Fong, P. Welinder, B. McGrew, J. Tobin, P. Abbeel, W. Zaremba, ``Hindsight Experience Replay,'' CoRR 2017
% \bibitem{b2} V. Pong, S. Gu, M. Dalal, S.Levine, ``Temporal Difference Models: Model-Free Deep {RL} for Model-Based Control,'' CoRR 2018
% \bibitem{b3} C. Florensa, D. Held, M. Wulfmeier, P. Abbeel, ``Reverse Curriculum Generation for Reinforcement Learning,'' CoRR 2017

% \end{thebibliography}

\end{document}